\pdfoutput=1

\documentclass[11pt]{article}

\usepackage[final]{ACL2023}

\usepackage{times}
\usepackage{latexsym}

\usepackage[T1]{fontenc}

\usepackage[utf8]{inputenc}

\usepackage{microtype}

\usepackage{inconsolata}

\usepackage{booktabs}%
\usepackage{multicol}
\usepackage{graphicx}
\usepackage{tcolorbox}

\usepackage{titlesec}
\titlespacing*{\section}{0pt}{0.3\baselineskip}{0\baselineskip}
\titlespacing*{\subsection}{0pt}{0.3\baselineskip}{0\baselineskip}
\titlespacing*{\subsubsection}{0pt}{0.3\baselineskip}{0\baselineskip}
\titlespacing*{\paragraph}{0pt}{0.1\baselineskip}{0\baselineskip}
\usepackage{enumitem}
\setlist[itemize]{itemsep=-3pt, topsep=1pt, partopsep=0pt}
\setlist[enumerate]{itemsep=0pt, topsep=0pt, partopsep=0pt}

\title{Classification of Human- and AI-Generated Texts\\ for English, French, German, and Spanish}

\author{Kristina Schaaff \and Tim Schlippe \and Lorenz Mindner \\
         IU International University of Applied Sciences, Germany.\\
         \texttt{kristina.schaaff@iu.org; tim.schlippe@iu.org}}

\begin{document}
\maketitle
\begin{abstract}

In this paper we analyze features to classify \textit{human-} and \textit{AI-generated} text for English, French, German and Spanish and compare them across languages. We investigate two scenarios: (1)~The detection of text generated by AI from scratch, and (2)~the detection of text rephrased by AI. 
For training and testing the classifiers in this multilingual setting, we created a new text corpus covering 10 topics for each language. 
For the detection of \textit{AI-generated} text, the combination of all proposed features performs best, indicating that our features are portable to other related languages: The F1-scores are close with 99\% for Spanish, 98\%~for English, 97\%~for German and 95\%~for French. For the detection of \textit{AI-rephrased} text, the systems with all features outperform systems with other features in many cases, but using only document features performs best for German (72\%) and Spanish (86\%) and only text vector features leads to best results for English (78\%).




\end{abstract}

\section{Introduction}

In recent years, chatbots have gained popularity and are now widely used in everyday life \cite{pelau2021makes}. These systems are designed to simulate \textit{human-}like conversations and provide assistance, information, and emotional support \cite{ dibitonto2018chatbot, arteaga2019design, falala2019owlie, adiwardana2020towards}. 
OpenAI's ChatGPT has emerged as one of the most commonly used tool for text generation \cite{taecharungroj2023can}. Within a short span of only five days after its release, over one million users registered~\cite{taecharungroj2023can}. The application scenarios are manifold, ranging from children seeking help with their homework to individuals seeking medical advice or companionship.

As the use of chatbots like ChatGPT becomes more prevalent in our daily lives, it is important to differentiate between \textit{human-generated} and \textit{AI-generated} text. 
As AI algorithms improve, detecting \textit{AI-generated} content accurately becomes increasingly challenging, posing issues such as plagiarism, fake news generation, and spamming. Thus, tools that can differentiate between \textit{human-} and \textit{AI-generated} content are crucial.

In \citet{mindner23}, we explored a large number of innovative features such as text objectivity, list lookup features, and error-based features for the detection of English (\textit{EN}) text generated by ChatGPT. However, in the current study, we extended this research to Spanish (\textit{ES}), German (\textit{DE}), and French (\textit{FR}). We selected these languages, as these are amongst the most frequently used languages in the world~\cite{Ethnologue2023}. 

Consequently, our contributions are as follows: 
\begin{itemize}
    \item We proved, that the features we investigated in \citet{mindner23}  can be successfully ported to other languages.
    \item We extended our \textit{Human-AI-Generated Text Corpus}\footnote{https://github.com/LorenzM97/human-AI-generatedTextCorpus}
    with \textit{FR}, \textit{DE} and \textit{ES} articles which cover 10~topics, providing a benchmark corpus for the detection of \textit{AI-generated texts} in \textit{EN}, \textit{FR}, \textit{DE} and \textit{ES}. 
    \item Our best systems  significantly outperform the state-of-the-art system for the detection of \textit{AI-generated} text ZeroGPT.
\end{itemize}


\section{Related Work}
\label{sec:relwork}

In the this section, we will describe the related work concerning ChatGPT and the classification of \textit{human-} and \textit{AI-generated} texts. 

\subsection{ChatGPT}

Since its release by OpenAI in late 2022, ChatGPT has revolutionized the field of AI~\cite{Mesko.2023} and several other generative AIs such as Google's Bard\footnote{https://bard.google.com} or Llama\footnote{https://ai.meta.com/llama}~\cite{touvron2023llama} have been released. 
Those tools are capable of generating text in response to user queries across a wide range of domains. Its successful implementation has been demonstrated in areas like education \cite{baidoo2023education}, medicine \cite{jeblick2022chatgpt}, and language translation \cite{jiao2023chatgpt}. ChatGPT is built on the Generative Pre-trained Transformers (GPT) language model and undergoes fine-tuning using reinforcement learning with human feedback. This approach allows ChatGPT to grasp the meaning and intention behind user prompts, enabling it to provide relevant and helpful responses. During the training process, a substantial amount of text data is incorporated to ensure the safety and accuracy of the generated text. While the quantity of training data has not been published, we know that the previous GPT-3 model, which is substantially larger than other language models such as BERT~\cite{devlin2018bert}, RoBERTa~\cite{liu2019roberta}, and T5~\cite{roberts2019t5}, was trained with 175 billion parameters and 499 billion crawled text tokens \cite{GPT3:2020}. 
Through extensive training on a diverse dataset, ChatGPT has acquired a sophisticated understanding of human language, allowing it to generate text that closely resembles that written by humans \cite{mitrovic2023chatgpt}.

\begin{table*}[!h]
\centering
\footnotesize
\begin{tabular}{@{}llllllllll@{}}
\toprule
    & \multicolumn{3}{c}{\textbf{Human}}     & \multicolumn{3}{c}{\textbf{AI-generated}}        & \multicolumn{3}{c}{\textbf{AI-rephrased}} \\ 
       \textbf{Language} & \textbf{P} & \textbf{S} & \textbf{W} & \textbf{P} & \textbf{S} & \textbf{W} & \textbf{P} & \textbf{S} & \textbf{W}  \\ \midrule
    EN & 415& 1.7k& 38.3k& 555& 1.4k& 27.6k& 255& 1.1k& 24.6k\\ 
    FR & 415& 1.2k& 31.0k& 524& 1.3k& 26.5k& 157& 0.8k& 18.7k \\
    DE & 335& 1.2k& 20.5k& 529& 1.4k& 22.9k& 256& 1.0k& 16.4k \\
    ES & 450& 1.4k& 38.0k& 514& 1.2k& 26.8k& 190& 0.8k& 18.9k \\
    \bottomrule
\end{tabular}
\caption{\textit{AI-Generated/Rephrased} Text \\(P = \#paragraphs, S = \#sentences, W = \#words).}
\label{table:statistics-basic}
\end{table*}

\subsection{Detecting \textit{Human-} and \textit{AI-Generated} Texts}
\label{sec:relworkclass}

Commercial tools and plagiarism apps, such as GPTZero~\cite{GPTZero}, ZeroGPT\footnote{https://www.zerogpt.com}, AI Content Detector\footnote{https://copyleaks.com/ai-content-detector},
and GPT-2 Output Detector\footnote{https://openai-openai-detector--mqlck.hf.space} \cite{DetectGPT}, have been developed to identify \textit{AI-generated} text. 
Furthermore, researchers are working on developing new corpora for this task and finding out which features and classifiers improve classification accuracy: For example, \cite{yu2023cheat} present a corpus of \textit{human-} and \textit{AI-generated} abstracts to investigate commercial and non-commercial systems---but only for \textit{EN}. Recent studies 
have explored various approaches to detect \textit{AI-generated} text, including XGBoost~\cite{Shijaku:2023}, decision trees~\cite{zaitsu2023distinguishing}, and transformer-based models ~\cite{mitrovic2023chatgpt,guo2023close}: \citet{mitrovic2023chatgpt} evaluated characteristics of \textit{AI-generated} text from \textit{EN} customer reviews and built a transformer-based classifier that achieved 79\%. 
\citet{zaitsu2023distinguishing} achieved 100\% accuracy in the detection of Japanese texts with decision trees combining stylometric features for Japanese such as bigrams, comma position, and function word rates. \citet{guo2023close} evaluated the characteristics of \textit{human-generated} and \textit{AI-generated} answers to questions in \textit{EN} and Chinese. They fine-tuned a RoBERTa model
on their texts and achieved 98.8\% F1-score on the \textit{EN} answers and 96.4\% F1-score on the Chinese answers. 
\citet{Shijaku:2023} addressed the detection of generated essays written in \textit{EN} and proposed an XGBoost model that achieved 98\% accuracy using features generated by TF-IDF and a set of hand-crafted features. \citet{soni2023comparing} analyzed \textit{human-} and \textit{AI-generated} text summarization and achieved 90\% accuracy using DistilBERT\footnote{https://huggingface.co/docs/transformers/model\_doc/distilbert}~\cite{DistilBERT}.
\citet{mindner23} explored features to detect \textit{AI-generated} and \textit{-rephrased} text for \textit{EN}. They report an F1-score of 96\% for \textit{AI-generated} text and 78\% for \textit{AI-rephrased} text on their text corpus which contains different topics. These F1-scores were even achieved when the AI was instructed to create the text in a way that a human would not recognize that it was generated by an AI.

To the best of our knowledge, we are the first to explore a large set of features and state-of-the-art classifiers across multiple languages with XGBoost, Random Forrest and MLP. We compare our results with two popular state-of-the-art tools that detect texts generated by AI: GPTZero and ZeroGPT. GPTZero is used by over 1~million people~\cite{GPTZero}, but its results are only reliable for \textit{EN} texts. Consequently, we also used ZeroGPT for comparison which is able to deal with other languages. As there is currently no text corpus available, which contains \textit{human-} and \textit{AI-generated} texts in multiple languages, we extended our \textit{Human-AI-Generated Text Corpus} to cover \textit{EN}, \textit{FR}, \textit{DE} and \textit{ES}.


\section{Our Human-AI-Generated Text Corpus}
\label{sec:datacorpus}

As mentioned in the previous section, we extended our \textit{Human-AI-Generated Text Corpus}~\cite{mindner23} to cover \textit{EN}, \textit{FR}, \textit{DE}, and \textit{ES}. 
In total, for each language we used 100~\textit{human-generated}, 100~\textit{AI-generated}, and  100~\textit{AI-rephrased} articles for our multilingual analysis which contain the following 10~topics: $biology$, $chemistry$, $geography$, $history$, $IT$, $music$, $politics$, $religion$, $sports$, and $visual arts$. 

The characteristics of our \textit{Human-AI-Generated Text Corpus} for the respective languages are summarized in Table \ref{table:statistics-basic}: 
\textit{EN} consistently has the highest counts across all categories and types of text. On the other hand, the counts for \textit{FR}, \textit{DE}, and \textit{ES} vary substantially depending on whether the text was \textit{human-generated}, \textit{AI-generated}, or \textit{AI-rephrased}. This illustrates how languages differ in the expression of information. 
The prompts which we used to receive the \textit{AI-generated} and \textit{AI-rephrased} texts are listed in Table~\ref{table:aigeneration}.

\begin{table}[ht]
\footnotesize
\begin{tabular}{@{}lp{6.5cm}@{}}
\toprule
\textbf{Lang.} & \textbf{Prompt} \\
\midrule
\multicolumn{2}{@{}l}{\textbf{Text Generation}}\\
EN & Generate a text on the following topic: <topic> \\
FR &Rédigez un texte sur le thème suivant: <topic> \\
DE & Erstelle einen Text zum folgenden Thema: <topic>  \\
ES & Genera un texto sobre el siguiente tema: <topic> \\
\midrule
\multicolumn{2}{@{}l}{\textbf{Text Rephrasing}}\\
EN & Rephrase the following text: <topic> \\
FR &Reformulez le texte suivant: <topic>\\
DE & Formuliere den folgenden Text um: <topic> \\
ES & Reformule el siguiente texto: <topic>\\
\bottomrule
\end{tabular}
\caption{Prompts used for Generation and Rephrasing}
\label{table:aigeneration}
\end{table}

\section{Our Features for the Classification of \textit{Human-} and \textit{AI-Generated} Texts}\label{sec:features}

As shown in Table~\ref{table:featuresummary}, we analyzed 37~features for their suitability to discriminate between \textit{human-} and \textit{AI-generated} text. More details of the features are given in~\citet{mindner23}. 


\begin{table*}[ht]
\centering
\footnotesize
\begin{tabular}{@{}lll@{}}
\toprule
\textbf{Category} & \textbf{Feature} & \textbf{Description} \\ \midrule
\textit{Perplexity} & $PPL_{mean}$ & mean PPL \\
 & $PPL_{max}$ & maximum PPL \\ \midrule
\textit{Semantic} & $sentiment_{polarity}$ & degree of positivity/negativity [-1,+1] \\
 & $sentiment_{subjectivity}$ & degree of subjectivity [0,+1] \\ \midrule
\textit{ListLookup} &$stopWord_{count}$ & number of stop words \\
&$discourseMarker_{count}$ & number of discourse markers \\
&$titleRepetition_{count}$ & absolute repetitions of title \\
&$titleRepetition_{relative}$ & relative repetitions of title \\ 
&$personalPronoun_{count}$ & absolute number of personal pronouns \\
&$personalPronoun_{relative}$ & relative number of personal pronouns \\
\midrule
\textit{Document}  & $wordsPerParagraph_{mean}$ & mean number of words per paragraph \\
&$wordsPerParagraph_{stdev}$ & stdev of $wordsPerParagraph$ \\ 
&$sentencesPerParagraph_{mean}$ & mean number of sentences per paragraph \\
&$sentencesPerParagraph_{stdev}$ & stdev of $sentencesPerParagraph$ \\
&$wordsPerSentence_{mean}$ & mean number of words per sentence \\
&$wordsPerSentence_{stdev}$ & stdev of $wordsPerSentence$ \\
&$uniqWordsPerSentence_{mean}$ & mean number of unique words per sentence \\
&$uniqWordsPerSentence_{stdev}$ & stdev of $uniqWordsPerSentence$ \\
&$words_{count}$ & number of running words \\
&$uniqWords_{count}$ & number of unique words \\
&$uniqWords_{relative}$ & relative number of unique words \\
&$paragraph_{count}$ & number of paragraphs \\
&$sentence_{count}$ & number of sentences \\
&$punctuation_{count}$ & number of punctuation marks \\
&$quotation_{count}$ & number of quotation marks \\
&$character_{count}$ & number of characters \\
&$uppercaseWords_{relative}$ & relative number of words in uppercase \\
&$POSPerSentence_{mean}$ & mean number of unique POS-tags/sentence \\
&$specialChar_{count}$ & number of special characters \\
\midrule
\textit{ErrorBased}& $grammarError_{count}$ & number of spelling/grammar errors \\
&$multiBlank_{count}$ & number of multiple blanks \\  \midrule
\textit{Readability} & $fleschReadingEase$ & Flesch Reading Ease score [0-100] \\
&$fleschKincaidGradeLevel$ & Readability as U.S. grade level [0-100] \\ \midrule
\textit{AIFeedback} & $AIFeedback$ & Ask AI if text was generated by AI \\ \midrule
\textit{TextVector} &\textit{TF-IDF} & 500-dim TF-IDF vector of 1-/2-grams \\
&\textit{Sentence-BERT} & mean Sentence-BERT vector \\
&\textit{Sentence-BERT-dist} & mean distance of Sentence-BERT vectors \\
\bottomrule
\end{tabular}
\caption{Summary of our Features for the Classification of Generated Texts.}
\label{table:featuresummary}
\end{table*}

\subsection{Perplexity-Based Features} 
Perplexity is a measure of how well a language model is able to predict a sequence of words. The lower the perplexity, the better a language model will perform to predict the next word in a sequence. As \textit{AI-generated} texts are usually based on statistical patterns and rules, they tend to be more repetitive and therefore have a lower perplexity than human generated texts.
The \textit{perplexity-based} features in our study are based on the findings by \citet{mindner23, gehrmann-etal-2019-gltr, mitrovic2023chatgpt, guo2023close}. 



For sentence tokenization, we use the Natural Language Toolkit (NLTK)\footnote{https://www.nltk.org}. Perplexity is calculated using \textit{evaluate package}\footnote{https://github.com/huggingface/evaluate} and GPT-2 using the respective models for \textit{EN}\footnote{https://huggingface.co/gpt2}, \textit{FR}\footnote{https://huggingface.co/dbddv01/gpt2-french-small}, \textit{DE}\footnote{https://huggingface.co/dbmdz/german-gpt2}, and \textit{ES}\footnote{https://huggingface.co/DeepESP/gpt2-spanish}.


\subsection{Semantic Features}
In our study, \textit{semantic} features refer to the properties of words or phrases used to represent their meanings. Previous studies successfully used these features for the differentiation between \textit{human-} and \textit{AI-generated} texts \cite{mitrovic2023chatgpt, guo2023close, mindner23}. 

Again, we use different Python packages for the respective languages: TextBlob's sentiment analysis for English\footnote{https://textblob.readthedocs.io/en/dev/quickstart.html}, \texttt{textblob-fr}\footnote{https://github.com/sloria/textblob-fr} for French, and \texttt{textblob-de}\footnote{https://textblob-de.readthedocs.io/en/latest/api\_reference .html\#module-textblob\_de.sentiments} for German. Due to the absence of a package that computes both, polarity and subjectivity, for \textit{ES} texts were translated these texts into \textit{EN} using Googletrans\footnote{https://github.com/ssut/py-googletrans}, despite potential information loss, because of its high BLEU score and proficiency in \textit{ES-EN} translation.

\subsection{List Lookup Features}
With our \textit{ListLookup} features, we analyze information about the word or character class, e.g., whether it is a stop word or a special character. These features have previously been used for this task by \citet{mindner23, Shijaku:2023, kumarage2023stylometric}.
For every language, we used ChatGPT to generate a list of all discourse markers as well as the personal pronouns. These lists were additionally evaluated by language experts. To count stop words, we use NLTK for the respective languages. 


\subsection{Document Features}
Our \textit{document} features are related to the content and structure of a document such as word frequencies, syntactic structures, and corpus statistics. These features have been successfully used by \cite{kumarage2023stylometric, Shijaku:2023, guo2023close,mitrovic2023chatgpt,zaitsu2023distinguishing,mindner23}.
To calculate \textit{sentence}- and \textit{word-related} features, the text is first divided into sentences and words using NLTK's \texttt{sent\_tokenize} and \texttt{word\_tokenize} functions.
For the features related to Part-of-speech (POS) in \textit{EN} texts, we use the NLTK function \texttt{pos\_tag}. As NLTK lacks POS tags for the other three languages, we use spaCy NLP library\footnote{https://github.com/explosion/spaCy}. For POS tags in \textit{DE} texts, we use \texttt{de\_core\_news\_sm}\footnote{https://spacy.io/models/de\#de\_core\_news\_sm}, for \textit{FR} texts, we use \texttt{fr\_core\_news\_sm}\footnote{https://spacy.io/models/fr\#fr\_core\_news\_sm}, and for \textit{ES} texts, \texttt{es\_core\_news\_sm}\footnote{https://spacy.io/models/es\#es\_core\_news\_sm}.


\subsection{Error Based Features}

This feature category introduced in \citet{mindner23} is based on errors in the text such as grammar and spelling mistakes. 

To count multiple blanks, we used regular expressions. Grammar and spelling errors are detected using the open-source tool \textit{LanguageTool}\footnote{https://github.com/jxmorris12/language\_tool\_python} which allows it to detect grammar errors in multiple languages. For the detection of \textit{DE} errors, the built-in class \texttt{LanguageToolPublicAPI(de-DE)} for querying the tool’s public servers is used. For the other languages, the tool’s remote server is applied using the function \texttt{Language-Tool(language)}.


\subsection{Readability Features}
\textit{Readability} features assess the readability level of texts as in \citet{mindner23,Shijaku:2023,Flesch1948ANR,Kincaid1975DerivationON}.


To derive Flesch Reading Ease and Flesch-Kincaid Grade Level we use \textit{Textstat}\footnote{https://github.com/textstat/textstat}. This Python library provides functions to calculate text statistics such as grade level, complexity, and readability.
Textstat supports calculating Flesch Reading Ease, and Flesch-Kincaid Grade Level for \textit{EN}, \textit{FR}, \textit{DE}, and \textit{ES} texts. However, it is important to note that these measures were originally developed for the specific structure of words, sentences, and syllables of \textit{EN}. Therefore, when applying these measures to texts in \textit{FR}, \textit{DE}, and \textit{ES}, the results may not be as representative as those for \textit{EN}.


\subsection{AI Feedback Features}
Our \textit{AI Feeback} features reflect, how an AI categorizes the text \cite{mindner23}. For this purpose, we use ChatGPT with the prompts in Table~\ref{table:aifeedback}. 


\begin{table}[ht]
\footnotesize
\begin{tabular}{@{}lp{6.5cm}@{}}
\toprule
\textbf{Lang.} & \textbf{Prompt} \\
\midrule
EN & Was the following text generated by ChatGPT?  \\
FR & Le texte suivant a-t-il été généré par ChatGPT? \\
DE & Wurde der folgende Text von ChatGPT generiert?  \\
ES & ¿El siguiente texto fue generado por ChatGPT?  \\
\bottomrule
\end{tabular}
\caption{Prompts used for AI Feedback.}
\label{table:aifeedback}
\end{table}

\begin{figure*}[htbp]
    \centering
    \includegraphics[width=1\linewidth]{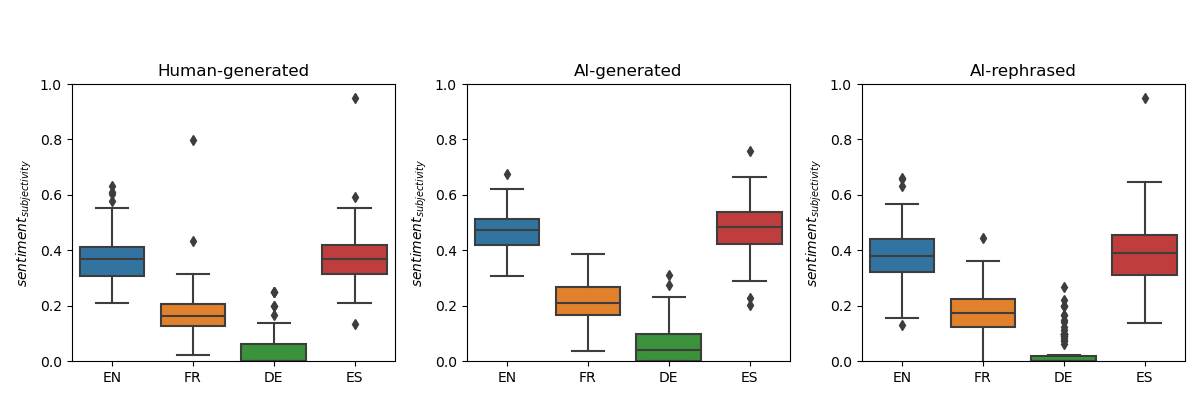}
    \caption{Distribution of $sentiment_{subjectivity}$}
    \vspace{09pt}
    \label{fig:subjectivity}
\end{figure*}




   





\subsection{Text Vector Features}
Our \textit{TextVector} features analyze semantic content of a text, identifying patterns and repetition \cite{mindner23, Shijaku:2023, solaiman2019release,reimers-gurevych-2019-sentence}.


For the features based on Sentence-BERT, we use the sentence-transformer model \texttt{distiluse- base-multilingual-cased-v2}\footnote{https://huggingface.co/sentence-transformers/distiluse-base-multilingual-cased-v2}, since it supports all the languages used in this research. In addition to the four languages in our experiments, it can be used for more than 50 languages, guaranteeing reliable results for possible future research.

\subsection{Summary of Our Analyzed Features}

Our 8 feature categories contain 37 features. While the \textit{AI feedback} category consists of one feature, the perplexity, semantic, error-based, and readability features each contain two features. The largest feature category are document features, which contains 19 different features. Table \ref{table:featuresummary} summarizes all the features that are part of our experiments.

\section{Experimental Setup}\label{sec:experimentalsetup}

In this section, we will describe our experiments with the different feature categories and three classification approaches: The two more traditional approaches XGBoost~\cite{Shijaku:2023} and random forest~(RF)~\cite{breiman2001random} as well as a neural network-based approach with multilayer perceptrons~(MLP)~\cite{MURTAGH1991183}.
As in other studies like \citet{guo2023close,kumarage2023stylometric,mitrovic2023chatgpt}, we evaluated the classification performance with accuracy (\textit{Acc}) and F1-score (\textit{F1}). 
First, we built \textit{text generation detection systems} which were trained, fine-tuned, and tested with our \textit{human-generated} and \textit{AI-generated} texts. Second, we implemented \textit{text rephrasing detection systems} which were trained, fine-tuned, and tested with our \textit{human-generated} and \textit{AI-rephrased} texts. 
To provide stable results, we used a 5-fold cross-validation, randomly dividing our corpus into 80\% training, 10\% validation, and 10\% unseen test set. The numbers in all tables are the average of the test set results. The best performances are highlighted in bold. 
As a baseline, we choose two popular state-of-the-art tools which detect texts generated by AI: GPTZero and ZeroGPT. GPTZero is used by over 1~million people~\cite{GPTZero}. 
However, we found that GPTZero's results were only reliable for \textit{EN} texts. Consequently, we used ZeroGPT 
as our baseline for \textit{FR}, \textit{DE} and \textit{ES}.

\begin{table*}[ht!]
\centering
\footnotesize
\begin{tabular}{@{}ll|rrrrrr|rrrrrr@{}}
\toprule
&& \multicolumn{6}{c}{\textbf{Generated}} & \multicolumn{6}{c}{\textbf{Rephrased}} \\
&& \multicolumn{2}{c}{\textbf{XGBoost}} & \multicolumn{2}{c}{\textbf{RF}} & \multicolumn{2}{c}{\textbf{MLP}} & \multicolumn{2}{c}{\textbf{XGBoost}} & \multicolumn{2}{c}{\textbf{RF}} & \multicolumn{2}{c}{\textbf{MLP}} \\
\textbf{Category} & \textbf{Lang} & \textbf{Acc} & \textbf{F1} & \textbf{Acc} & \textbf{F1} & \textbf{Acc} & \textbf{F1}& \textbf{Acc} & \textbf{F1} & \textbf{Acc} & \textbf{F1} & \textbf{Acc} & \textbf{F1}\\
\toprule
\textit{Perplexity}&EN& 83.0 & 82.2 & 87.0 & 85.3 & 82.0 & 82.1 & 52.0 & 48.7 & 55.0 & 54.6 & 56.0 & 63.2 \\
&FR& 62.0 & 60.3 & 69.0 & 66.8 & 68.0 & 69.0 & 50.0 & 50.2 & 53.0 & 44.2 & 56.0 & 58.8 \\
&DE& 74.0 & 74.0 & 76.0 & 76.1 & 81.0 & 80.6 & 53.0 & 53.6 & 61.0 & 60.4 & 56.0 & 62.7 \\
&ES& 82.0 & 82.3 & 83.0 & 82.4 & 82.0 & 83.6 & 56.0 & 55.4 & 63.0 & 63.7 & 62.0 & 67.3 \\
\midrule
\textit{Semantic}&EN& 72.0 & 72.9 & 75.0 & 75.6 & 73.0 & 72.3 & 66.0 & 64.4 & 66.0 & 64.3 & 52.0 & 54.3 \\
&FR& 61.0 & 55.8 & 67.0 & 65.6 & 63.0 & 59.4 & 55.0 & 48.2 & 57.0 & 50.0 & 51.0 & 52.9 \\
&DE& 64.0 & 58.3 & 64.0 & 59.8 & 63.0 & 63.3 & 56.0 & 59.9 & 54.0 & 54.4 & 62.0 & 60.1 \\
&ES& 72.0 & 69.9 & 75.0 & 73.8 & 76.0 & 75.7 & 58.0 & 56.1 & 58.0 & 52.4 & 53.0 & 56.3 \\
\midrule
\textit{ListLookup}&EN& 72.0 & 72.1 & 79.0 & 78.5 & 71.0 & 67.8 & 72.0 & 73.9 & 67.0 & 67.5 & 69.0 & 70.3 \\
&FR& 72.0 & 73.0 & 76.0 & 76.7 & 67.0 & 62.9 & 66.0 & 62.6 & 65.0 & 65.5 & 64.0 & 63.2 \\
&DE& 74.0 & 75.8 & 79.0 & 77.8 & 72.0 & 74.1 & 57.0 & 59.1 & 58.0 & 59.2 & 50.0 & 52.0 \\
&ES& 78.0 & 79.6 & 82.0 & 84.1 & 73.0 & 76.8 & 75.0 & 75.2 & 80.0 & 81.3 & 77.0 & 78.4 \\
\midrule
\textit{Document}&EN& 91.0 & 91.6 & 92.0 & 92.6 & 87.0 & 86.0 & 70.0 & 69.6 & 71.0 & 70.8 & 78.0 & 76.1 \\
&FR& 94.0 & 94.2 & 91.0 & 90.8 & 92.0\textbf{ }& 92.2 & 86.0 & 85.3 & 84.0 & 80.8 & 81.0 & 81.2 \\
&DE& 87.0 & 87.2 & 90.0 & 89.6 & 88.0 & 88.0 & \textbf{72.0} & \textbf{71.9} & 67.0 & 66.7 & 71.0 & 71.3 \\
&ES& 96.0 & 96.2 & 98.0 & 98.1 & 87.0 & 88.5 & 84.0 & 83.4 & 83.0 & 82.0 & \textbf{86.0} & \textbf{86.4} \\
\midrule
\textit{ErrorBased}&EN& 55.0 & 61.7 & 55.0 & 61.7 & 56.0 & 63.9 & 62.0 & 68.0 & 62.0 & 68.0 & 62.0 & 68.0 \\
&FR& 62.0 & 64.2 & 63.0 & 67.2 & 61.0 & 65.5 & 53.0 & 56.0 & 56.0 & 58.9 & 56.0 & 59.7 \\
&DE& 67.0 & 67.1 & 67.0 & 67.1 & 67.0 & 69.8 & 62.0 & 61.9 & 62.0 & 63.5 & 56.0 & 50.7 \\
&ES& 70.0 & 71.2 & 71.0 & 71.9 & 71.0 & 74.6 & 59.0 & 56.8 & 61.0 & 56.3 & 64.0 & 65.2 \\
\midrule
\textit{Readability}&EN& 60.0 & 56.3 & 63.0 & 59.3 & 60.0 & 56.8 & 54.0 & 51.1 & 54.0 & 47.8 & 50.0 & 50.2 \\
&FR& 61.0 & 64.7 & 62.0 & 66.0 & 65.0 & 67.4 & 59.0 & 58.3 & 60.0 & 60.6 & 52.0 & 31.6 \\
&DE& 57.0 & 53.5 & 53.0 & 51.5 & 57.0 & 53.6 & 48.0 & 41.9 & 45.0 & 39.1 & 45.0 & 44.9 \\
&ES& 74.0 & 73.7 & 74.0 & 72.1 & 69.0 & 66.6 & 54.0 & 49.1 & 61.0 & 50.7 & 56.0 & 52.5 \\
\midrule
\textit{AIFeedback}&EN& 62.0 & 67.1 & 62.0 & 67.1 & 62.0 & 68.1 & 52.0 & 50.9 & 50.0 & 39.8 & 45.0 & 30.1 \\
&FR& 52.0 & 24.2 & 52.0 & 24.2 & 48.0 & 37.2 & 42.0 & 33.6 & 42.0 & 33.6 & 55.0 & 53.4 \\
&DE& 49.0 & 46.1 & 47.0 & 35.0 & 50.0 & 43.4 & 52.0 & 61.8 & 52.0 & 61.8 & 50.0 & 54.3 \\
&ES& 52.0 & 7.3  & 52.0 & 7.3  & 52.0 & 20.6 & 50.0 & 0.0  & 52.0 & 7.3  & 49.0 & 25.7 \\
\midrule
\textit{TextVector}&EN& 90.0 & 89.9 & 95.0 & 94.9 & 83.0 & 81.7 & \textbf{79.0} & \textbf{78.2} & 75.0 & 71.0 & 69.0 & 65.1 \\
&FR& 94.0 & 94.1 & 93.0 & 93.0 & 85.0 & 85.4 & 77.0 & 77.3 & 75.0 & 75.2 & 68.0 & 64.2 \\
&DE& 87.0 & 87.0 & 94.0 & 94.0 & 90.0 & 90.8 & 68.0 & 67.5 & 72.0 & 67.3 & 72.0 & 71.7 \\
&ES& 84.0 & 84.5 & 91.0 & 89.5 & 81.0 & 76.6 & 76.0 & 74.0 & 76.0 & 73.6 & 68.0 & 64.4 \\
\midrule
\textit{All}&EN& 90.0 & 90.9 & \textbf{98.0} & \textbf{98.0} & 87.0 & 87.8 & 77.0 & 77.6 & 71.0 & 69.8 & 72.0 & 71.9 \\
&FR& 94.0 & 94.4 & \textbf{95.0} & \textbf{95.0} & 88.0 & 89.2 & \textbf{89.0} & \textbf{87.9} & 86.0 & 84.2 & 74.0 & 66.4 \\
&DE& 94.0 & 93.8 & \textbf{97.0} & \textbf{96.9} & 87.0 & 86.6 & 70.0 & 71.6 & 71.0 & 68.3 & 70.0 & 71.6 \\
&ES& 94.0 & 94.4 & \textbf{99.0} & \textbf{99.0} & 90.0 & 90.2 & 83.0 & 82.2 & 83.0 & 82.9 & 78.0 & 76.1 \\
\bottomrule
\end{tabular}
\caption{Results for the Detection of \textit{EN} \textit{FR}, \textit{DE} and \textit{ES} \textit{AI-generated} and AI-Rephrased Texts.}
\label{tab:results:AI-rephrased}
\end{table*}

\section{Results}\label{sec:experiments}

Table~\ref{tab:results:AI-rephrased} lists \textit{Acc} and \textit{F1} for
detecting \textit{AI-generated} and \textit{-rephrased} texts in  \textit{EN}, \textit{FR}, \textit{DE}, and \textit{ES}. 
For each language classifiers trained on \textit{AI-generated} texts achieve better performances compared to classifiers trained on \textit{AI-rephrased} texts.

\subsection{Results of Single Feature Categories}

As shown in Figure~\ref{fig:subjectivity} using the example of $sentiment_{subjectivity}$, the distribution of feature values can differ depending on whether the text is \textit{human-generated}, \textit{AI-generated} or \textit{AI-rephrased} and depending on the language.
$sentiment_{subjectivity}$ denotes objectivity (low values) or subjectivity (high values) of a text. Average $sentiment_{subjectivity}$ values tend to be higher for \textit{AI-generated} text than for \textit{human-generated} and \textit{AI-rephrased} text. 
In general, \textit{DE} texts are the most objective texts---be it \textit{human-} or \textit{AI-generated}---while \textit{EN} and \textit{ES} are more subjective. Moreover, \textit{AI-generated} texts tend to be more subjective than \textit{AI-rephrased} texts for our languages.







\vspace{-0.1cm}

\subsubsection{English}

\vspace{-0.1cm}

\paragraph{Text Generation Detection\ \ \ }

The results for \textit{EN} in Table~\ref{tab:results:AI-rephrased} indicate that the system that combines all features (\textit{All}) in an RF performs best (\textit{Acc}=98.0\%, \textit{F1}=98.0\%). The 2nd-best system is the MLP system that uses \textit{Document} features (\textit{Acc}=95.0\%, \textit{F1}=94.9\%). The RF system that uses \textit{TextVector} features results in a similar performance (\textit{Acc}=95.0\%, \textit{F1}=94.9\%). The worst-performing system is the XGBoost system that uses the \textit{ErrorBased} features (\textit{Acc}=55.0\%, \textit{F1}=61.7\%). Compared to GPTZero ($Acc_{GPTZero}$=76.0\%,  $F1_{GPTZero}$=78.9\%), most of our systems perform better. Our best system with all features (\textit{All}) outperforms GPTZero by 28.9\% relative in \textit{Acc} and 24.2\% relative in \textit{F1}. ZeroGPT reaches 78.0\% $Acc_{ZeroGPT}$ and 81.8\% $F1_{ZeroGPT}$. Thus, our best system performs 25.6\% relatively better in \textit{Acc}, and 19.8\% relatively better in \textit{F1}. 

\paragraph{Text Rephrasing Detection\ \ \ }

The performances for the \textit{EN} \textit{text rephrasing detection systems} are worse than the \textit{text generation detection systems} for all feature categories except \textit{ErrorBased} (\textit{Acc}=62.0\%, \textit{F1}=68.0\%). The best-performing system is the XGBoost system that uses \textit{TextVector} features (\textit{Acc}=79.0\%, \textit{F1}=78.2\%), followed by the MLP system that uses \textit{Document} features (\textit{Acc}=78.0\%, \textit{F1}=76.1\%). The worst-performing system is the MLP system that uses the \textit{AIFeedback} feature. All our \textit{text rephrasing detection systems} were able to outperform GPTZero ($Acc_{GPTZero}$=43.0\% and $F1_{GPTZero}$=27.8\%). Our the best-performing \textit{TextVector} feature system even outperforms GPTZero by 83.7\% relative in \textit{Acc} and even 159.8\% relative in \textit{F1}. ZeroGPT reaches 49.0\% $Acc_{ZeroGPT}$ and 43.9\% $F1_{ZeroGPT}$. Thus, \textit{Document} outperforms it by 61.2\% relative in \textit{Acc} and 81.5\% relative in \textit{F1}.

\subsubsection{French}

\paragraph{Text Generation Detection\ \ \ }

The results for \textit{FR} in Table~\ref{tab:results:AI-rephrased} demonstrate that the system that combines all features (\textit{All}) in an RF performs best (\textit{Acc}=95.0\%, \textit{F1}=95.0\%). 
The 2nd-best system is the XGBoost system that uses \textit{Document} features (\textit{Acc}=86.0\%, \textit{F1}=85.3\%), followed by the XGBoost system that uses \textit{TextVector} features (\textit{Acc}=77.0\%, \textit{F1}=77.3\%). The worst-performing systems are those that use the \textit{AIFeedback} feature. Our best \textit{FR} system with all features (\textit{All}) outperforms ZeroGPT ($Acc_{ZeroGPT}$=62.0, $F1_{ZeroGPT}$)=72.6\%) by 53.2\%  relative in \textit{Acc} and 30.9\% relative in \textit{F1}. 

\paragraph{Text Rephrasing Detection\ \ \ }

The performances for the \textit{FR} \textit{text rephrasing detection systems} are worse than the \textit{text generation detection systems} for all feature categories except the MLP system that uses the \textit{AIFeedback} feature (\textit{Acc}=55.0\%, \textit{F1}=53.4\%). The best-performing system is the system that that combines all features (\textit{All}) in an XGBoost (\textit{Acc}=89.0\%, \textit{F1}=87.9\%), followed by the XGBoost system that uses \textit{Document} features (\textit{Acc}=86.0\%, \textit{F1}=85.3\%) and the XGBoost system that uses \textit{TextVector} features (\textit{Acc}=77.0\%, \textit{F1}=77.3\%). The worst-performing systems are again those that use the \textit{AIFeedback} feature. Our best \textit{FR} system with all features~(\textit{All}) outperforms ZeroGPT ($Acc_{ZeroGPT}$=57.0, $F1_{ZeroGPT}$)=67.4\%) by 56.1\%  relative in \textit{Acc} and 30.4\% relative in \textit{F1}. 

\subsubsection{German}

\paragraph{Text Generation Detection\ \ \ }

The results for \textit{DE} in Table~\ref{tab:results:AI-rephrased} indicate that the system that combines all features (\textit{All}) in an RF performs best (\textit{Acc}=97.0\%, \textit{F1}=96.9\%). 
The 2nd-best system is the RF system that uses \textit{TextVector} features (\textit{Acc}=94.0\%, \textit{F1}=94.0\%), followed by the RF system that uses \textit{Document} features (\textit{Acc}=90.0\%, \textit{F1}=89.6\%). As for the previous languages, the worst-performing systems are those that use the \textit{AIFeedback} feature. Our best \textit{FR} system with all features (\textit{All}) outperforms ZeroGPT ($Acc_{ZeroGPT}$=65.0, $F1_{ZeroGPT}$)=70.9\%) by 49.2\%  relative in \textit{Acc} and 36.7\% relative in \textit{F1}. 

\paragraph{Text Rephrasing Detection\ \ \ }

The performances for the \textit{DE} \textit{text rephrasing detection systems} are worse than the \textit{text generation detection systems} for all feature categories except the systems that use the \textit{AIFeedback} features. The best-performing system is the XGBoost system that that uses the \textit{Document} features (\textit{Acc}=72.0\%, \textit{F1}=71.9\%), followed by the MLP system that uses \textit{TextVector} features (\textit{Acc}=72.0\%, \textit{F1}=71.7\%). The worst-performing systems are those that use the \textit{Readability} feature. Our best \textit{DE} system with the \textit{Document} features outperforms ZeroGPT ($Acc_{ZeroGPT}$=48.0, $F1_{ZeroGPT}$=49.5\%) by 45.5\% relative in \textit{Acc} and 45.3\% relative in \textit{F1}. 

\subsubsection{Spanish}

\paragraph{Text Generation Detection\ \ \ }

The results for \textit{ES} in Table~\ref{tab:results:AI-rephrased} show that the system that combines all features (\textit{All}) in an RF performs best (\textit{Acc}=99.0\%, \textit{F1}=99.0\%). 
The 2nd-best system is the RF system that uses \textit{Document} features (\textit{Acc}=98.0\%, \textit{F1}=89.1\%), followed by the RF system that uses \textit{TextVector} features (\textit{Acc}=91.0\%, \textit{F1}=89.5\%) and the RF system that uses \textit{ListLookup} features (\textit{Acc}=82.0\%, \textit{F1}=84.1\%). As for the previous languages, the worst-performing systems are those that use the \textit{AIFeedback} feature. The \textit{F1} of 7.3\% is so poor since the feature classifies the text as \textit{AI-generated} text in almost all cases. Our best \textit{ES} system with all features (\textit{All}) outperforms ZeroGPT ($Acc_{ZeroGPT}$=60.0, $F1_{ZeroGPT}$)=71.5\%) by 65.0\%  relative in \textit{Acc} and 38.5\% relative in \textit{F1}. 

\paragraph{Text Rephrasing Detection\ \ \ }
The performances for the \textit{ES} \textit{text rephrasing detection systems} are worse than the \textit{text generation detection systems} for all feature categories. 
The best-performing system is the RF system that uses the \textit{Document} features (\textit{Acc}=86.0\%, \textit{F1}=86.4\%). The 2nd best system is the system that combines all features (\textit{All}) in an RF (\textit{Acc}=83.0\%, \textit{F1}=82.9\%), followed by the RF system that uses the \textit{ListLookup} features (\textit{Acc}=80.0\%, \textit{F1}=81.3\%). 
The worst-performing systems are those that use the \textit{AIFeedback} feature. The \textit{F1} of 0\% and 7.3\% are so poor since the feature classifies the text as AI generated text in almost all cases. Our best \textit{ES} system with the \textit{Document} features outperforms ZeroGPT ($Acc_{ZeroGPT}$=52.0, $F1_{ZeroGPT}$=63.7\%) by 65.4\% relative in \textit{Acc} and 25.6\% relative in \textit{F1}.

\subsubsection{Combination of All Features}

As shown in Table~\ref{tab:results:AI-rephrased}, the best performances for the text generation detection systems are achieved using a combination of all features (\textit{All}). Looking at the systems which use all features, the \textit{Acc} for the \textit{AI-generated} \textit{FR} and \textit{DE} texts is similar with 97.0\%, while the \textit{Acc} for the \textit{AI-generated} \textit{EN} texts is 98.0\%. The best \textit{F1} for the \textit{AI-generated} \textit{DE} classifier is 96.9\%. Thus, it is slightly worse than the classifiers trained on our \textit{EN} and \textit{FR} texts which achieved 98.0\% and 97.1\%, respectively. The best classifier trained on the \textit{AI-generated} \textit{ES} texts achieved slightly better performances, with 99.0\% \textit{Acc} and 99.0\% \textit{F1}. Comparing the performances of the systems trained on the \textit{AI-generated} texts, it can be summarized that the classifiers deliver comparable performances across the languages.

The performances of the systems which use all features (\textit{All}) vary more for the \textit{AI-rephrased} texts across the languages. While the best \textit{EN} classifier reaches 79.0\% \textit{Acc} on the \textit{AI-rephrased} texts, the best \textit{FR} classifier achieves 89.0\% \textit{Acc} on the \textit{AI-rephrased} texts. The \textit{AI-rephrased} detection system for \textit{DE} only achieves 72.0\% \textit{Acc}. Compared to the best \textit{DE} text rephrasing detection system, the \textit{FR} system is 23.6\% relatively better in~\textit{Acc}. 
The \textit{Acc} for the \textit{ES} text rephrasing detection system is 1\% worse than the \textit{FR} system. For \textit{F1}, comparable conclusions can be drawn across the languages. Thus, our investigated features do not deliver comparable performances for the detection of \textit{AI-rephrased} texts across the evaluated languages.

\section{Conclusion and Future Work}\label{sec:conclusion}


In this paper, we investigated features to classify whether text is written by a human, generated by AI from scratch or rephrased by AI. 
We conducted a comparative analysis of the classification across the languages of \textit{EN}, \textit{FR}, \textit{DE}, and \textit{ES}, assessing the performance of these features in their respective linguistic contexts. To train and test classifiers which use the features, we extended the Human-AI-Generated Text Corpus~\cite{mindner23}---our new text corpus, which covers 10~different topics for each of the four languages. For \textit{AI-generated text}, our classifier performed best when combining all features, meaning that there are no substantial differences for features across languages. Therefore, we conclude, that the same feature set could also be used for other languages from the same language families. The accuracies are close with 99\% for \textit{ES}, 98\%~for \textit{EN}, 97\%~for \textit{DE} and 95\%~for \textit{FR}.
In contrast to that, for the detection of {AI-rephrased} text, the systems with all features outperformed systems with other features in many cases. For \textit{DE} (72\%) and \textit{ES} (86\%)  we achieved the best results using only document features while for \textit{EN} the text vector features yielded the best results (79\%). 
 
Although our results indicate that the same feature set could be applied to other languages within the same familie, future work could investigate the applicability of these features across further language families. This would help in understanding the robustness of our method across a more diverse set of languages. Moreover, our corpus currently covers 10 different topics for each language. Extending the corpus to include more topics, and possibly considering different domains and genres, may help in generalizing the findings and making the system more robust. Finally, experimenting with different machine learning architectures such as transformer models could potentially lead to further optimizations.




\section*{Ethics Statement}

The collected corpus is made freely available to the community. It is based on Wikipedia and news texts. 
The research was conducted transparently, 
free from bias and in compliance with applicable laws and regulations. The use of AI models and data 
is intended to foster a deeper understanding of AI-generated content, with the goal of promoting responsible use and technological innovation.

\bibliography{custom}
\bibliographystyle{acl_natbib}

\end{document}